\documentclass[a4paper]{article}

\usepackage{INTERSPEECH2018}
\usepackage{multirow}

\title {Gated Recurrent Unit Based Acoustic Modeling with Future Context}
\name{Jie Li$^1$, Xiaorui Wang$^1$, Yuanyuan Zhao$^2$, Yan Li$^1$}
\address{
  $^1$Kwai, Beijing, P.R. China \\
  $^2$Institute of Automation, Chinese Academy of Sciences, Beijing, P.R.China}
\email{\{lijie03, wangxiaorui, liyan\}@kuaishou.com, yyzhao5231@ia.ac.cn}

\begin{document}

\maketitle

\begin{abstract}
The use of future contextual information is typically shown to be helpful for acoustic modeling. However, for the recurrent neural network (RNN), it's not so easy to model the future temporal context effectively, meanwhile keep lower model latency. In this paper, we attempt to design a RNN acoustic model that being capable of utilizing the future context effectively and directly, with the model latency and computation cost as low as possible. The proposed model is based on the minimal gated recurrent unit (mGRU) with an input projection layer inserted in it. Two context modules, \emph{temporal encoding} and \emph{temporal convolution}, are specifically designed for this architecture to model the future context. Experimental results on the Switchboard task and an internal Mandarin ASR task show that, the proposed model performs much better than long short-term memory (LSTM) and mGRU models, whereas enables online decoding with a maximum latency of 170 ms. This model even outperforms a very strong baseline, TDNN-LSTM, with smaller model latency and almost half less parameters. 
\end{abstract}

\section{Introduction}


It is typically shown to be beneficial for acoustic modeling to make full use of the future contextual information.  In the literature, there are variety of methods to realize this idea for different model architectures. For feed-forward neural network (FFNN), this context is usually provided by splicing a fixed set of future frames in the input representation\cite{DongYu}. It also exists other approaches relating modifying FFNN model structures. The authors in \cite{FSMN1,FSMN2} proposed a model called feedforward sequential memory networks (FSMN), which is a standard FFNN equipped with some learnable memory blocks in the hidden layers to encode the long context information into a fixed-size representation. The time delay neural network (TDNN) \cite{SPL2,SPL17} is another FFNN architecture which has been shown to be effective in modeling long range dependencies through temporal convolution over context. 

As for unidirectional recurrent neural network (RNN), this is usually accomplished using a delayed prediction of the output labels\cite{ASRU6}. However, this method only provides quite limited modeling power of future context, as shown in \cite{SPL}. While for bidirectional RNN, this is accomplished by processing the data in the backward direction using a separate RNN layer \cite{SPL4,SPL5,SPL6}. Although the bidirectional versions have been shown to outperform the unidirectional ones with a large margin \cite{SPL7,SPL8}, the latency of bidirectional models is significantly larger, making them unsuitable for online speech recognition. To overcome this limitation, chunk based training and decoding schemes such as context-sensitive-chunk (CSC) \cite{SPL10, Chen2016A} and latency-controlled (LC) BLSTM \cite{SPL7, SPL12} have been investigated. However, the model latency is still quite high, since in all these online variants, inference is restricted to chunk-level increments to amortize the computation cost of backward RNN.  For example, the decoding latency of LC-BLSTM in \cite{SPL12} is about 600 ms, which is the sum of chunk size $N_c$ and future context frames $N_r$. To overcome the shortcomings of the chunk-based methods, 
Peddinti \emph{et al.} \cite{SPL} proposed the use of temporal convolution, in the form of TDNN layers, for modeling the future temporal context while affording inference with frame-level increments. The proposed model is called TDNN-LSTM, and is designed by interleaving of temporal convolution (TDNN layers) with unidirectional long short-term memory (LSTM)  \cite{ASRU1,ASRU2,ASRU3,ASRU4} layers. This model was shown to outperform bidirectional LSTM in two automatic speech recognition (ASR) tasks, while enabling online decoding with a maximum latency of 200 ms \cite{SPL}.



However, TDNN-LSTM's ability to model the future context comes from the TDNN part, whereas the LSTM itself is incapable of utilizing the future information effectively. 
In this paper, we attempt to design a RNN acoustic model that can model the future context effectively and directly, without the dependence on extra layers, for instance, TDNN layers. In addition, the model latency and computation cost should be as low as possible. 

With this purpose, we choose to use the minimal gated recurrent unit (mGRU) \cite{Bengio} as our base RNN model in this work. mGRU is a revised version of GRU \cite{Bengio28,Bengio29} and contains only one multiplicative gate, making the computational cost of mGRU much smaller than GRU and vanilla LSTM \cite{ASRU4}. Based on mGRU, we propose to insert a linear input projection layer to mGRU, getting a model called mGRUIP. The inserted linear projection layer compresses the input vector and hidden state vector simultaneously. Since the size of this layer is much smaller than cell number, mGRUIP contains much less parameters than mGRU. In addition to this, there are two other advantages of the input projection layer. The first one is that inserting this layer is beneficial to the ASR performance. Our experiments on a 309-hour Switchboard task show that mGRUIP outperforms mGRU significantly. This finding is consistent with that in LSTM with input projection layer (LSTMIP) \cite{ASRU}. 

The second (also the most important) advantage is that this input projection forms a bottleneck in the recurrent layer, making it possible to design a module on it, that can utilize the future context information effectively, meanwhile without significantly increasing the model size. In this work, we design two kinds of context modules specifically for mGRUIP, making it capable of modeling future temporal context effectively and directly. The first module is referred to as \emph{temporal encoding}, in which one mGRUIP layer is equipped with a context block to encode the future context information into a fixed-size representation, similar with FSMN. Temporal encoding is performed at the input projection layer, making the increase of computation cost quite small. The second module borrows the idea from TDNN, and is called \emph{temporal convolution} as the transforms in it are tied across time steps. In temporal convolution, future context information from several frames is spliced together and compressed by the input projection layer. Thanks to the small dimensionality of the projection, temporal convolution brings quite limited additional parameters. In this work, these two context modules are shown to be quite effective on two ASR tasks, while maintaining low latency (170 ms) online decoding. It is shown that compared with LSTM and mGRU, mGRUIP with temporal convolution provides more than 13\% relative WER reduction on the full Switchboard Hub5'00 test set, while on our 1400-hour internal Mandarin ASR task, the relative gain is 13\% to 24\% for different test sets. What's more, the proposed model outperforms TDNN-LSTM with smaller decoding latency and almost half less parameters.

This paper is organized as follows. Section 2 describes the model architecture of GRU and its variants, including the proposed mGRUIP and the two context modules. The related work is introduced in Section 3. We report our experimental results on two ASR tasks in Section 4 and conclude this work in Section 5.

\section{Model Architecures}
In this section, we will first make a brief introduction to the model structure of GRU and mGRU. Then the proposed mGRUIP and two context modules will be introduced in detail. 

\subsection {GRU}
The GRU model is defined by the following equations (the layer index $l$ has been omitted for simplicity):
\begin{eqnarray}
  && z_t = \sigma(W_zx_t + U_zh_{t-1} + b_z) \\
  && r_t = \sigma(W_rx_t + U_rh_{t-1} + b_r) \\
  && \widetilde{h_t} = \tanh(W_hx_t + U_h(h_{t-1}*r_t) + b_h) \\
  && h_t = z_t*h_{t-1} + (1-z_t)*\widetilde{h_t}
\end{eqnarray}

In particular, $z_t$ and $r_t$ are vectors corresponding to the update and reset gates respectively, where $*$ denotes element-wise multiplication. The activations of both gates are element-wise logistic sigmoid functions $\sigma(\cdot)$, constraining the values of $z_t$ and $r_t$ ranging from 0 to 1. $h_t$ represents the output state vector for the current time frame $t$, while $\widetilde{h_t}$ is the candidate state obtained with a hyperbolic tangent. The network is fed by the current input vector $x_t$ (speech features or output vector of previous layer), and the parameters of the model are $W_z$, $W_r$, $W_h$ (the feed-forward connections), $U_z$, $U_r$, $U_h$ (the recurrent weights), and the bias vectors $b_z$, $b_r$, $b_h$.

\subsection {mGRU}
mGRU, short for minimal GRU, is a revised version of the GRU described above. It is proposed by \cite{Bengio} and contains two modifications: removing the reset gate and replacing the hyperbolic tangent function with ReLU activation. Thus it leads to the following update equations:
\begin{eqnarray}
  && z_t = \sigma(W_zx_t + U_zh_{t-1} + b_z) \\
  && \widetilde{h_t} = {\rm ReLU}({\rm BN}(W_hx_t + U_hh_{t-1}) + b_h) \\
  && h_t = z_t*h_{t-1} + (1-z_t)*\widetilde{h_t}
\end{eqnarray}
where BN means batch normalization.

\subsection {mGRUIP}
In this work, a novel model called mGRUIP is proposed by inserting a linear input projection layer into mGRU. In mGRUIP, the output state vector $h_t$ is calculated from the input vector $x_t$ by the following equations:
\begin{eqnarray}
  && v_t = W_v[x_t ; h_{t-1}] \\
  && z_t = \sigma(W_zv_t + b_z) \\
  && \widetilde{h_t} = {\rm ReLU}({\rm BN}(W_hv_t) + b_h) \\
  && h_t = z_t*h_{t-1} + (1-z_t)*\widetilde{h_t}
\end{eqnarray}

In mGRUIP, the current input vector $x_t$ and the previous output state vector $h_{t-1}$, are concatenated together and compressed into a lower dimensional projected vector $v_t$ by weight matrices $W_v$. Then the update gate activation $z_t$ and the candidate state vector $\widetilde{h_t}$ are calculated based on the projected vector $v_t$.

mGRUIP can reduce the parameters of mGRU significantly. The total number of parameters in a standard mGRU network, ignoring the biases, can be computed as follows:
\begin{equation*}
 N_{mGRU} = n_i {\times n_c} {\times 2} + n_c {\times n_c} {\times 2}
\end{equation*}
where $n_c$ is the number of hidden neurons, $n_i$ the number of input units, and $N_{mGRU}$ is the total parameter number of mGRU. While for mGRUIP, this value becomes:
\begin{equation*}
 N_{mGRUIP} = (n_i + n_c) {\times n_p} + n_p {\times n_c} {\times 2}
\end{equation*}
where $n_p$ is the number of units in the input projection layer. Assuming $n_c$ equal with $n_i$, the ratio of these two numbers is:
\begin{equation*}
\frac{N_{mGRUIP}}{N_{mGRU}} =  \frac {n_p}{n_c}
\end{equation*}
In a typical configuration we can set $n_c=1024$ and $n_p=512$, hence the parameters of mGRUIP is just half of mGRU, making the computation quite efficient. Despite this, our experiments on Switchboard task show that mGRUIP outperforms mGRU with the same number of neurons, i.e., $n_c$. What's more, increasing $n_c$ while decreasing $n_p$ can further enlarge the gains.
 
\subsection {mGRUIP with Context Module}
The input projection layer forms a bottleneck in mGRUIP, making it easier to utilize the future context effectively, in the meantime keep the increase of model size acceptable. In this paper, two kinds of context module, namely \emph{temporal encoding} and \emph{temporal convolution}, are specifically designed for mGRUIP.

\subsubsection {mGRUIP with Temporal Encoding}
In temporal encoding, context information from several future frames are encoded into a fixed-size representation at the input projection layer. Thus equation (8) in a standard mGRUIP now becomes:
\begin{eqnarray}
  && v_t^l = {W_v^l}[x_t^l ; h_{t-1}^l] + \sum_{i=1}^K f(v_{t+s\times i}^{l-1})
\end{eqnarray}
where the last summation part in equation (12) stands for temporal encoding. In particular, $v_{t+s\times i}^{l-1}$ is the input projection vector of layer $l-1$ from the $(t+s\times i)$th frame. $s\ge 1$ is the step stride and $K$ is the order of future context. $f(\cdot)$ denotes the transform function applied to $v_t^{l-1}$. In this work, we tried 3 forms: identity ($f(x)=x$), scale ($f(x)=m*x$) and affine transform ($f(x)=Wx$). Preliminary results show that the identity function gives slightly better performance than the other two forms. Thus we choose $f(x)=x$ for the rest of this paper. It should be noted that in this case, temporal encoding brings no additional parameters for mGRUIP.

\subsubsection {mGRUIP with Temporal Convolution}
Temporal encoding uses the projection vector of lower layer ($v_{t+s\times i}^{l-1}$) to represent the future context, while in temporal convolution, the future information is extracted from the output state vector of lower layer and then compressed by the input projection. Equation (8) now becomes:
\begin{eqnarray}
   && v_t^l = {W_v^l}[x_t^l ; h_{t-1}^l] + {W_p^l}[h_{t+s\times i}^{l-1};\cdots;h_{t+s\times K}^{l-1}]
\end{eqnarray}
where the last part represents temporal convolution. In particular, $h_{t+s\times i}^{l-1}$ is the output state vector of layer $l-1$ on the $(t+s\times i)$th frame. Same as temporal encoding, $s$ is the step stride and $K$ is the context order. According to this equation, $h_{t+s\times i}^{l-1}$ from $K$ future frames are spliced together and projected to a lower dimensional space by matrix $W_p^l$. Assuming the number of hidden neurons in layer $l-1$ is $n_c$, temporal convolution brings $K\times n_c \times n_p$ additional parameters. However, since the value of $n_p$ is usually quite small and we generally splice no more than two frames ($K \le 2$), the increase of the model size is limited and acceptable. 

\section {Related Work}
The authors in \cite{ASRU} proposed to insert an input projection layer to vanilla LSTM to reduce the computation cost. In this work, we tried this idea on mGRU\cite{Bengio}, getting a model called mGRUIP, which is shown to be more effective and more efficient than mGRU.

TDNN-LSTM \cite{SPL} is one of the most powerful acoustic model that can utilize future context effectively while has relatively low model latency. However, the ability of modeling the future temporal context comes from TDNN and has nothing to do with the LSTM layers. In this work, thanks to the input projection layer, we empower the mGRUIP to be capable of modeling the future context effectively and directly, by equipping it with one of the two proposed context modules, \emph{temporal encoding} and \emph{temporal convolution}. These two modules borrows the ideas from FSMN \cite{FSMN1, FSMN2} and TDNN  \cite{SPL2,SPL17} respectively. The difference is that, FSMN and TDNN belong to FFNN, therefore both of them need to model the future context as well as the past information to capture the long-term dependencies. Whereas the two proposed context modules are placed in a RNN layer, and they only need to focus on the future context, leaving the history to be modeled by recurrent connections.

Row convolution \cite{Deep2}, which encodes future context by applying a context-independent weight matrix, is another method to model the future context for RNN. The idea is similar with the two proposed context modules. However, row convolution in \cite{Deep2} is only placed  above all recurrent layers. While in this work, we place context modules in all hidden layers (except the first one). This layer-wise context expansion makes the higher layers having the ability to learn wider temporal  relationships than lower layers. What's more, the objective function is also different: connectionist temporal classification (CTC) \cite{CTC} in \cite{Deep2} while lattice-free MMI (LF-MMI) \cite{SPL16} in this work.



\section {Experiments}
In this section, we evaluate the effectiveness and efficiency of the proposed mGRUIP on two ASR tasks. The first one is the 309-hour Switchboard conversational telephone speech task, and the second one is an internal Mandarin voice input task with 1400-hour training data. All the models in this paper are trained LF-MMI objective function computed on 33Hz outputs \cite{SPL16}.

\subsection {Switchboard ASR Task}
The training data set consists of 309-hour Switchboard-I training data. Evaluation is performed in terms of word error rate (WER) on the full Switchboard Hub5'00 test set, consisting of two subsets: Switchboard (SWB) and CallHome (CHE). The experimental setup follows \cite{SPL16}. We use the speed-perturbation technique \cite{Chain16} for 3-fold data augmentation, and iVectors to perform instantaneous adaptation of the neural network \cite{Chain17}. WER results are reported after 4-gram LM rescoring of lattices generated using a trigram LM. For details about the model training, the reader is directed to \cite{SPL16}.

\subsubsection {Baseline Models}
Two baseline models, LSTM and mGRU, are trained for this task. Both of them contain 5 hidden layers, and the cell number for each layer is 1024. For LSTM, we add a recurrent projection layer on top of the memory blocks with a dimension of 512, compressing the cell output from 1024 to 512 dimension. For mGRU, to reduce the parameters of softmax output matrix, we insert a 512-dimensional linear bottleneck layer between the last hidden layer and the softmax layer. Both models are trained with an output delay of 50 ms. The input feature to both models at time step $t$ is a spliced version from frame $t-2$ through $t+2$. Therefore, they both have a model latency of 70 ms. Following \cite{SPL}, we use a mixed frame rate (MFR) across layers. In particular, the first hidden layer is operated at 100Hz frame rate while the rest of higher layers use a frame rate of 33Hz.

\subsubsection {mGRUIP}
To evaluate the effectiveness of the proposed mGRUIP, we train two models containing 5 layers, mGRUIP-A and mGRUIP-B, with different architectures. In mGRUIP-A, each hidden layer consists 1024 cells ($n_c=1024$, same as the baseline models), and the input projection layer has 512 units ($n_p=512$). While for mGRUIP-B, the cell number is 2560 and the projection dimension is 256. The training configurations are kept same as the baseline models.

\begin{table}[th]\small
  \caption{Performance comparison of LSTM, mGRU and mGRUIP on Switchboard task.}
  \label{tab:swb1}
  \centering
  \begin {tabular}{c|c|ccc}
  \toprule
  \multirow{2}*{Model} & {\#Param} & \multicolumn{3}{c}{WER (\%)} \\
   & (M) & SWB & CHM & Total \\
  \midrule
  LSTM & 19.7 & 10.3 & 20.7 & 15.6 \\
  mGRU  & 22.1 & 10.2 & 20.6 & 15.5 \\
  mGRUIP-A & 13.1 & 9.8 & 19.0 & 14.5 \\
  mGRUIP-B & 16.2 & \textbf{9.7}  &  \textbf{18.8}   & \textbf{14.3} \\
  \bottomrule
  \end{tabular}
\end{table}

The performance of the two mGRUIP models and two baseline models is shown in Table 1. We can see that, for these two baseline models, mGRU has more parameters and performs slightly better than LSTM. The proposed model mGRUIP-A contains much less parameters than the baseline mGRU (13.1M vs. 22.1M), but performs significantly better on the full test set (14.5 vs. 15.5). This means that the input projection layer can not only reduce the parameter of mGRU, but also being beneficial to the performance. It is also shown that mGRUIP-B outperforms mGRUIP-A, meaning that we can improve the ASR performance by increasing the cell number, meanwhile without significantly increasing the model size by reducing the projection dimension in mGRUIP. Compared with mGRU, mGRUIP-B provides 7.7\% relative WER reduction on the full test set whereas using 5.9M less parameters. In the following experiments, we will set $n_c=2560$ and $n_p=256$ for the mGRUIP related models.

\subsubsection {mGRUIP with Context Modules}
It's obvious that temporal encoding and temporal convolution can utilize more future context information by increasing $K$ and $s$ in equation (12) and (13). However, this will lead to the increase of model latency and model parameters (for temporal convolution). In this work, we did a lot of experiments and found the most cost-effective settings for these two context modules are as follows:
\begin{table}[th]
  \caption{The most cost-effective settings for two context modules.}
  \label{tab:swb2}
  \centering
  \begin {tabular}{c|cccc}
  \toprule
  Layer & $l=2$ & $l=3$ & $l=4$ & $l=5$ \\
  \midrule
  $K\times s$ & $1\times 1$ & $1\times 3$ & $1 \times 3$ & $1 \times 3$ \\
  \bottomrule
  \end{tabular}
\end{table}

As shown in Table 2, all the four higher mGRUIP layers (except the first one) are equipped with context modules. The context order $K$ for all of them is 1, and the step stride $s$ is 3 for the highest three layers while being 1 for the second hidden layer ($l=2$), making the operating frame rates same as the baselines. After equipped context modules with this setting, the latency of mGRUIP is increased from 70 ms to 170 ms. Table 3 shows the performance of mGRUIP with these two context modules. We also train a TDNN-LSTM model following \cite{SPL}, and the results are shown in the second line of Table 3.
\begin{table}[th]\footnotesize
  \caption{Performance comparison of LSTM, mGRU and mGRUIP on Switchboard task.}
  \label{tab:swb3}
  \centering
  \begin {tabular}{c|c|c|ccc}
  \toprule
  \multirow{2}*{Model} & {\#Param} &{Latency} & \multicolumn{3}{c}{WER (\%)} \\
   & (M) & (ms) & SWB & CHM & Total \\
  \midrule
  LSTM & 19.7 & 70 & 10.3 & 20.7 & 15.6 \\
  TDNN-LSTM & 34.8 & 200 & \textbf{9.0} & 19.7 & 14.4 \\
  \midrule
  mGRUIP-B & 16.2 & 70 & 9.7 & 18.8 & 14.3 \\
  +Ctx Encd & 16.2 & 170 & 9.5 & 18.0 & 13.8 \\
  +Ctx Conv & 18.7 & 170 & 9.2 & \textbf{17.8} & \textbf{13.5} \\
  \midrule
  \scriptsize MFR-BLSTM\cite{SPL} & - & 2020 & 9.0 & - & 13.6 \\
  \scriptsize TDNN-BLSTM-C\cite{SPL} & - & 2130 & 9.0 & - & 13.8 \\
  \bottomrule
  \end{tabular}
\end{table}

Several observations can be found in Table 3. First, both of the two context modules can improve the ASR performance of mGRUIP. Temporal convolution is more powerful than temporal encoding, while brings some additional parameters. Second, compared to LSTM, mGRUIP-B equipped with temporal convolution provides 13.5\% relative WER reduction, with a fraction of the cost of 100 ms additional model latency. Third, mGRUIP-B with temporal convolution is more effective than TDNN-LSTM on the full test set (13.5 vs. 14.4), with smaller model latency and much less parameters (18.7M vs. 34.8M). What's more, compared with the two most powerful models in \cite{SPL} (the last two lines of Table 3), the proposed model outperforms them on the full set with much smaller model latency (170 ms vs. 2000 ms). 

\subsection {Internal Mandarin ASR Task}
The second task is an internal Mandarin ASR task, of which the training set contains 1400 hours mobile recording data. The performance is evaluated on five public-available test sets, including three clean and two noisy ones. The three clean sets:
\begin{itemize}
\item AiShell\_dev: the development set of the released corpus AiShell-1\cite{Aishell}, containing 14326 utterances.
\item AiShell\_test: the test set of the released corpus AiShell-1, containing 7176 utterances.
\item THCHS-30\_Clean: the clean test set of THCHS-30 database\cite{thchs30}, containing 2496 utterances. 
\end{itemize}

The two noisy test sets are:
\begin{itemize}
\item THCHS-30\_Car: the corrupted version of THCHS-30\_Clean by car noise, the noise level is 0db. 
\item THCHS-30\_Cafe: the corrupted version of THCHS-30\_Clean by cafeteria noise, the noise level is 0db. 
\end{itemize}

Three ASR systems are built for this task: LSTM, TDNN-LSTM and mGRUIP-B with temporal convolution. The model architectures and the training configurations are all the same as Switchboard task. Results are shown in Table 4.
\begin{table}[th]\footnotesize
  \caption{Performance of different models on internal Mandarin ASR task.}
  \label{tab:man1}
  \centering
  \begin {tabular}{c|c|c|cc}
  \toprule
  \multirow{2}*{Test} & \multirow{2}*{LSTM} & \multirow{2}*{TDNN-LSTM} & \multicolumn{2}{c}{mGRUIP} \\
   & & & CER(\%) & CERR \\
  \midrule
  AiShell\_dev& 5.39 & 4.81 & 4.66 & 13.5\% \\
  AiShell\_test & 6.62 & 5.98 & 5.71 & 13.8\%  \\
  THCHS-30\_Clean & 11.93 & 10.97 & 10.38 & 13.0\% \\
  \midrule
  THCHS-30\_Car & 12.69 & 11.38 & 10.77 & 15.1\% \\
  THCHS-30\_Cafe & 53.19 & 44.20 & 40.26 & 24.3\% \\
  \bottomrule
  \end{tabular}
\end{table}

CERR column in Table 4 means the relative CER reduction of mGRUIP over LSTM. It's shown that mGRUIP performs much better than the baseline LSTM model on this task. On the three clean test sets, the CERR is about 13\%, and the gain is even larger on the two very noisy sets, from 15\% to 24\%. 

\section{Conclusions}

The aim of this paper is to design a RNN acoustic model that being capable of utilizing the future context effectively and directly, with the model latency and computation cost as low as possible. To achieve this goal, we choose the minimal GRU as our base model and propose to insert an input projection layer into it to further reduce the parameters. To model the future context effectively, we design two kinds of context modules, \emph{temporal encoding} and \emph{temporal convolution}, specifically for this architecture. Experimental results on the Switchboard task and an internal Mandarin ASR task show that, the proposed model performs much better than LSTM and mGRU models, whereas enables online decoding with a latency of 170 ms. This model even outperforms a very strong baseline, TDNN-LSTM, with smaller model latency and almost half less parameters.



\bibliographystyle{IEEEtran}

\bibliography{mybib}


\end{document}